\documentclass[10pt,twocolumn,letterpaper]{article}

\usepackage[pagenumbers]{iccv} 

\newcommand\blfootnote[1]{%
  \begingroup
  \renewcommand\thefootnote{}\footnote{#1}%
  \addtocounter{footnote}{-1}%
  \endgroup
}

\definecolor{iccvblue}{rgb}{0.21,0.49,0.74}
\usepackage[pagebackref,breaklinks,colorlinks,allcolors=iccvblue]{hyperref}
\usepackage{graphicx}
\usepackage{siunitx}
\usepackage{multirow}
\usepackage{tabularx,booktabs}
\usepackage{array}
\usepackage[table,xcdraw]{xcolor}
\usepackage{caption}
\usepackage{arydshln}

\def\paperID{*****} 
\def\confName{ICCV}
\def\confYear{2025}

\title{From Neural Activity to Computation: Biological Reservoirs for Pattern Recognition in Digit Classification}

\author{Ludovico Iannello$^{*1}$, Luca Ciampi$^{*1}$, Fabrizio Tonelli$^3$, Gabriele Lagani$^1$, Lucio Maria Calcagnile$^2$,\\Federico Cremisi$^3$, Angelo Di Garbo$^2$, Giuseppe Amato$^1$\\
$^1$ISTI-CNR, Italy\quad$^2$IBF-CNR, Italy\quad$^3$Bio@SNS, Italy\\
$^*${\tt\small ludovico.iannello@isti.cnr.it}\quad\quad$^*${\tt\small luca.ciampi@isti.cnr.it}}

\begin{document}
\maketitle
\begin{abstract}
In this paper, we present a biologically grounded approach to reservoir computing (RC), in which a network of cultured biological neurons serves as the reservoir substrate. This system, referred to as biological reservoir computing (BRC), replaces artificial recurrent units with the spontaneous and evoked activity of living neurons. A multi-electrode array (MEA) enables simultaneous stimulation and readout across multiple sites: inputs are delivered through a subset of electrodes, while the remaining ones capture the resulting neural responses, mapping input patterns into a high-dimensional biological feature space.
We evaluate the system through a case study on digit classification using a custom dataset. Input images are encoded and delivered to the biological reservoir via electrical stimulation, and the corresponding neural activity is used to train a simple linear classifier. To contextualize the performance of the biological system, we also include a comparison with a standard artificial reservoir trained on the same task. The results indicate that the biological reservoir can effectively support classification, highlighting its potential as a viable and interpretable computational substrate.
We believe this work contributes to the broader effort of integrating biological principles into machine learning and aligns with the goals of human-inspired vision by exploring how living neural systems can inform the design of efficient and biologically plausible models.
\blfootnote{$^*$Equal contribution.}
\end{abstract}    
\section{Introduction}
\label{sec:intro}

Reservoir computing (RC)~\cite{DBLP:journals/csr/LukoseviciusJ09} is a machine learning framework that transforms input data through the dynamics of a high-dimensional system. This transformation often results in representations that are more easily separable by simple classifiers, making RC a practical and efficient approach for tasks such as classification and regression (e.g., time-series analysis~\cite{DBLP:journals/tnn/BianchiSLJ21} and speech recognition~\cite{DBLP:journals/ficn/YonemuraK24}). Its effectiveness has been demonstrated across a range of implementations. One of the most prominent RC models is the echo state network (ESN)~\cite{jaeger2001,DBLP:journals/corr/abs-1712-04323}, in which a large pool of randomly connected recurrent units transforms input into a high-dimensional nonlinear representation that facilitates downstream learning. A related paradigm is the liquid state machine (LSM)~\cite{DBLP:journals/neco/MaassNM02,DBLP:journals/tnn/ZhangLJC15}, which employs spiking neurons to capture rich temporal dynamics. 



In this work, we propose an instantiation of the reservoir computing (RC) paradigm where the reservoir is not simulated but physically realized through a living network of cultured neurons. We refer to this approach as \textit{biological reservoir computing} (BRC). Unlike conventional RC systems that rely on artificial units, our method leverages the intrinsic dynamics of a biological neural culture to project input stimuli into a high-dimensional feature space. This setup offers a unique opportunity to explore the computational potential of real neural tissue, while also contributing to the broader goal of integrating biological principles into machine learning. Indeed, this biologically grounded approach not only provides a natural substrate for computation, but also aligns with neuroscientific evidence suggesting that transient neural dynamics play a key role in information processing~\cite{https://doi.org/10.1111/j.1460-9568.2007.05976.x}. Moreover, by offloading part of the computation to a physical system, BRC may offer advantages in terms of energy efficiency—an increasingly relevant concern in modern AI systems~\cite{DBLP:conf/aiia/BadarVSGMIAZ21,JAVED2010907}.

Specifically, the BRC system is built upon a high-density multi-electrode array (MEA)~\cite{bonifazi}, which enables both electrical stimulation and high-resolution recording of neural activity. Neurons are derived from stem cells using established differentiation protocols~\cite{Chambers_2009, PMID:33139941}, and once matured, they form spontaneously active networks. Input patterns are encoded as spatially distributed stimulation sequences and delivered to the culture via the MEA. The resulting neural responses are recorded and used to construct feature vectors for downstream classification. 
To evaluate the ability of our BRC system to generate discriminative feature representations, we conducted an experimental study focused on a single yet challenging pattern recognition task: the classification of ten digit-like spatial input patterns (from 0 to 9). Each pattern was defined by a distinct spatial configuration of electrical stimulation sites on a multi-electrode array (MEA). The evoked neural responses were recorded and used as features for a simple readout layer to classify the stimulus category.

Despite the inherent variability in biological responses due to noise, spontaneous activity, and differences across stimulation sessions and biological replicates, our system achieved promising levels of accuracy. These results demonstrate that cultured neuronal networks, even without considering plasticity or learning mechanisms, can serve as effective reservoirs that transform static spatial inputs into rich, high-dimensional representations suitable for downstream classification tasks. 
This manuscript is structured as follows. Sec.~\ref{sec:rel_work} surveys existing literature on reservoir computing and biological computation. Sec.~\ref{sec:method} details the experimental methodology, including MEA interfacing and stimulation protocols. In Sec.~\ref{sec:exp}, we present and analyze our experimental setup and results. Finally, Sec~\ref{sec:conclusions} shapes the conclusion and outlines directions for future research.

\begin{figure*}[t]
    \centerline{\includegraphics[width=0.95 \linewidth]{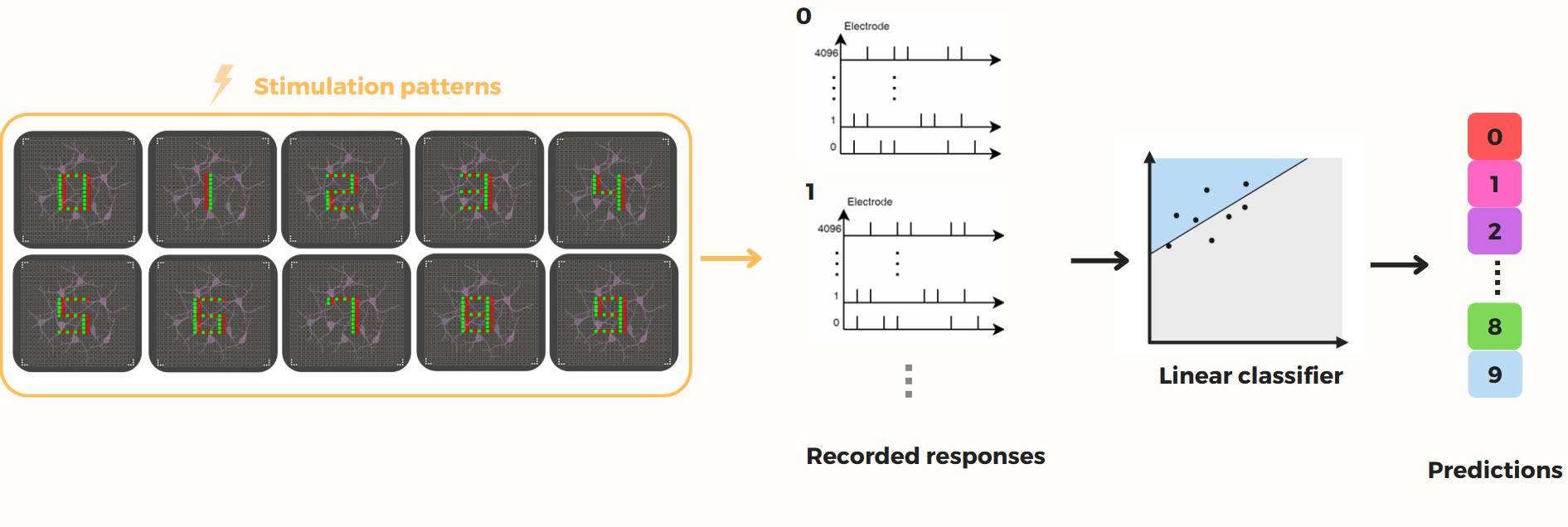}}
    \caption{\textbf{Framework of our Biological Reservoir Computing (BRC) paradigm.} In this approach, a multi-electrode array (MEA) functions as a bidirectional interface, enabling both the stimulation of and the recording from a cultured biological neural network. Discrete inputs are encoded by selectively activating specific subsets of MEA electrodes, which deliver targeted stimuli to the network. The evoked spiking responses are captured via a separate set of electrodes and transformed into high-dimensional vectors that encode the input within a latent computational space. Due to the inherent complexity and rich dynamics of the biological reservoir, this transformation is intrinsically nonlinear. Subsequently, a linear classifier is trained to infer the category of the original input from its corresponding latent representation.}
    \label{fig:teaser}
\end{figure*}

\section{Related Works} 
\label{sec:rel_work}

A central topic in deep learning (DL) research concerns the biological plausibility of existing computational frameworks. Numerous studies have critically examined the extent to which contemporary neural architectures reflect the structural and functional complexity of biological neural systems~\cite{DBLP:journals/ficn/MarblestoneWK16,Hassabis_2017,f3a024ff8e474a1ea4e798aaa5536830,DBLP:conf/atal/Tenenbaum18}. In light of this, there has been a growing movement toward the development of biologically inspired alternatives aimed at expanding the capabilities of machine learning and cognitive modeling~\cite{DBLP:journals/corr/abs-2307-16236,DBLP:journals/corr/abs-2307-16235}. Efforts in this direction have led to a wide array of models that attempt to bridge the gap between artificial and biological intelligence, either by implementing more biologically grounded neural computations~\cite{DBLP:journals/ficn/Diehl015,DBLP:journals/ficn/FerreMT18,DBLP:conf/ner/LaganiMFGCPCA21,DBLP:journals/cogcom/SunCCS23} or by refining the underlying synaptic and learning mechanisms~\cite{DBLP:journals/pnas/KrotovH19,DBLP:journals/nn/IllingGB19,DBLP:conf/mod/LaganiFGA21,DBLP:journals/nca/LaganiFGA22,DBLP:conf/cbmi/LaganiBGFGA22,DBLP:conf/sisap/LaganiGFA22,DBLP:conf/iclr/JourneRGM23,DBLP:journals/ijon/LaganiFGFA24,hebbian_eccv_workshop,DBLP:journals/corr/abs-2412-03192}.

Among these biologically inspired paradigms, reservoir computing (RC)~\cite{DBLP:journals/csr/LukoseviciusJ09,DBLP:journals/cogcom/ScardapaneBBM17} has attracted significant attention for its capacity to emulate complex neural dynamics. Two prominent subclasses within RC are echo state networks (ESNs)~\cite{DBLP:journals/corr/abs-1712-04323,jaeger2001} and liquid state machines (LSMs)~\cite{DBLP:journals/neco/MaassNM02,DBLP:journals/tnn/ZhangLJC15}, which differ in their computational mechanisms and biological plausibility. While LSMs typically employ spiking neuron models~\cite{DBLP:books/cu/GerstnerK02,PhysRevE.48.1483} to generate rich temporal dynamics, ESNs rely on a large, recurrent network of randomly connected continuous-valued units to embed input signals into a high-dimensional space conducive to downstream classification tasks. Although the recurrent architecture is initialized stochastically, specific strategies are employed to ensure that the network operates within a stable and computationally useful dynamic regime~\cite{DBLP:journals/corr/abs-1712-04323,DBLP:conf/inista/SarliGM20}. Several enhancements to the ESN framework have been proposed to increase its biological plausibility. For instance, models incorporating synaptic plasticity mechanisms, such as spike-timing-dependent plasticity (STDP)~\cite{Song_2000,DBLP:books/cu/GerstnerK02}, aim to more closely emulate biological learning processes~\cite{DBLP:conf/icpr/WangL16}. More recently, gating mechanisms have been introduced to improve the recall of long-term dependencies in nonlinear recurrent systems~\cite{DBLP:conf/inista/SarliGM20}. These RC-inspired techniques have demonstrated promising results across various application domains, including speech recognition~\cite{DBLP:journals/tnn/ZhangLJC15} and continual learning~\cite{DBLP:conf/esann/CossuBCGL21}.

Building upon this foundation, the present work introduces an extension to the RC paradigm by incorporating biological neural networks as computational reservoirs, i.e., a biological reservoir computing (BRC) wherein cultured neuronal populations serve as the dynamic substrate for computation. Although prior studies have explored the use of multi-electrode array (MEA) devices to interface with biological neurons~\cite{Shahaf8782,DBLP:journals/tbe/RuaroBT05,DBLP:journals/ijon/FerrandezLPF13,DBLP:journals/ploscb/IsomuraKJ15,GOEL2016320,DBLP:journals/ploscb/PastoreMGM18,KAGAN20223952}, only a few works have investigated the potential of employing biological neural networks as computational reservoirs~\cite{cai2023brain, iannello2025neurons}. Notably, Cai et al.~\cite{cai2023brain} employed brain organoids as biological reservoirs, demonstrating their capacity for speech recognition by leveraging the rich temporal dynamics inherent in such 3D biological structures. Their approach capitalizes on time-dependent neural activity patterns to encode and process input sequences. In contrast, our study focuses on spatially distributed stimulation patterns delivered to 2D cultured biological neural networks, rather than 3D organoids. Specifically, we employ spatially patterned electrical stimulation across a microelectrode array (MEA) without relying on temporal sequencing, aiming instead at static pattern recognition. This represents a distinct paradigm, emphasizing spatial encoding of input patterns over their temporal evolution.
Similarly, our preliminary work~\cite{iannello2025neurons} explored the feasibility of stimulating 2D cultured biological neural networks using a limited set of spatial patterns and network configurations. However,~\cite{iannello2025neurons} was primarily exploratory, involving a small number of stimuli and simpler classification tasks applied to a single biological network. By systematically investigating a broader range of spatially distributed stimulation patterns and evaluating their effects across three independent biological replicates (BRs), the present study constitutes one of the first comprehensive efforts to assess the feasibility of using cultured biological neural networks as functional reservoirs within a reservoir computing (RC) framework for static pattern recognition.
This approach not only bridges biological neural substrates and machine learning architectures, but also opens new avenues for leveraging the intrinsic properties of biological networks -- such as energy efficiency and complex nonlinear dynamics -- to advance neuromorphic computation.
\section{Biological Reservoir Computing} 
\label{sec:method}

\begin{figure}[t]
    \centerline{\includegraphics[width=0.6\linewidth]{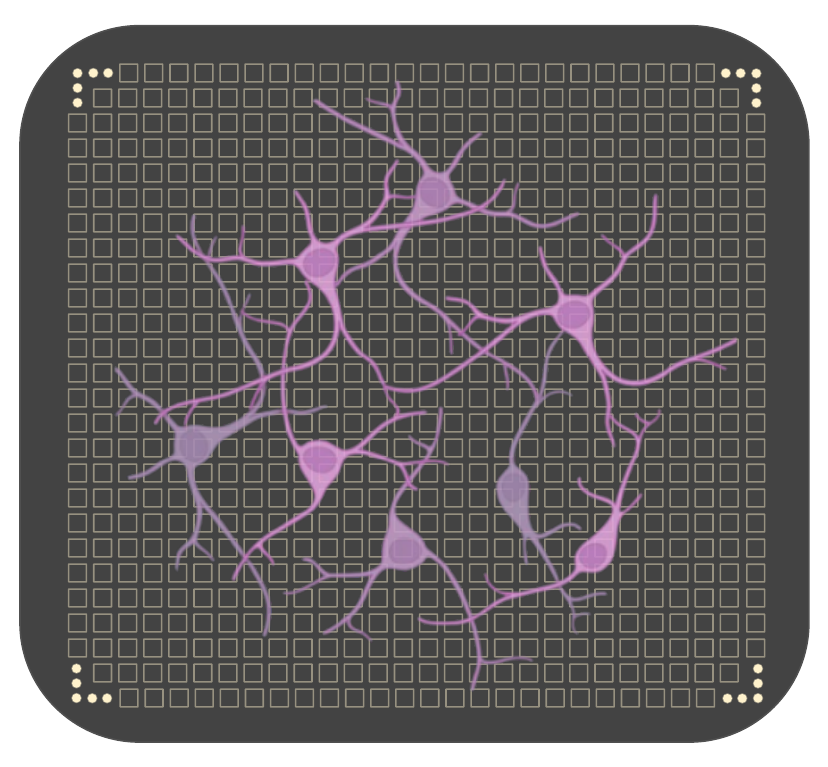}}
    \caption{\textbf{Illustration of a cultured biological network plated on a MEA device.} Each square represents a single MEA electrode. Input patterns are mapped onto the MEA by assigning elements of the patterns to specific electrodes. Electrical pulses are delivered based on the corresponding input intensities, and the evoked activity of the network is recorded via the remaining electrodes. The resulting spiking responses are used to construct a high-dimensional representation of the input in the feature space of the biological reservoir. Neuron scale in the figure is not to scale and is adjusted for visibility.}
    \label{fig:mea}
\end{figure}

A cultured network of biological neurons may be regarded as a complex system of randomly interconnected processing elements capable of generating rich, nonlinear dynamical behavior. When interfaced with a multi-electrode array (MEA) -- a bidirectional platform equipped for both electrical stimulation and electrophysiological recording (see ~\cref{fig:mea}) -- this neural substrate can be functionally harnessed to receive structured input stimuli and produce measurable responses.

In our reservoir computing paradigm, we define a direct correspondence between individual input samples and specific subsets of MEA electrodes. For instance, when an image is used as input, its pixel grid of dimensions $H \times W$ is projected onto an $H \times W$ subset of the electrode array. Each pixel is uniquely mapped to a designated electrode, and the pixel intensity modulates the stimulation parameters delivered to the underlying neurons. Following the stimulation phase, we collect the evoked spiking activity through the remaining electrodes on the MEA. These recorded spike trains are then aggregated into a high-dimensional feature vector, effectively embedding the input into the latent space defined by the biological reservoir.

In this configuration, the cultured neural network operates as a biologically instantiated feature extractor, transforming structured inputs into complex, nonlinear, high-dimensional representations~\cite{PMID:33139941,DBLP:journals/tbe/RuaroBT05}.

To quantitatively assess the quality and discriminative power of these biologically derived representations, we employ a single-layer perceptron as a downstream classifier. The full computational pipeline -- including stimulation protocol and classification framework -- is illustrated in ~\cref{fig:teaser}. The following sections detail the procedures for stimulation, feature extraction, training, and performance evaluation of the perceptron within our experimental setup.

\subsection{Stimulation Protocol}
Interfacing with a biological neural network involves several technical challenges, particularly with respect to the calibration of stimulation parameters such as pulse amplitude, frequency, and duration. These parameters must be finely tuned to ensure that stimulation elicits consistent and robust neural responses while maintaining the integrity of the hardware. In this work, we systematically investigate and define a set of optimized stimulation protocols designed to activate the cultured network effectively and support reliable electrophysiological recordings.

To deliver stimulation, specific electrodes within the MEA are designated as active stimulation sites. We used a bipolar stimulation, where each electrode pair is configured with opposing polarities, serving as the positive and negative poles. Each stimulus is delivered as a rectangular biphasic waveform with a defined amplitude $\si{\ampere}$ (expressed in $\si{\micro\ampere}$) and phase durations $\delta_+$ and $\delta_-$ (in $\si{\micro\second}$) corresponding to the positive and negative components, respectively. This approach enhanced the current balance between the positive and negative poles, contributing to the long-term stability and integrity of the electrodes. Since the MEA distributes current across multiple electrode pairs, the amplitude is interpreted on a per-pair basis. When determining these stimulation parameters, particular care is taken to balance signal efficacy with hardware preservation: the stimuli must be sufficiently strong to reliably evoke spiking activity, yet not so intense or frequent as to cause electrode degradation—a risk observed during initial trials involving high-current or high-frequency configurations. 

To mitigate any potential bias in the network's response due to order effects or temporal dependencies, all stimulation patterns are presented to the neuronal culture in a randomized sequence. A fixed inter-stimulus interval of $T = \SI{10}{\second}$ is enforced between successive inputs. This delay allows the network to return to a baseline or resting state before the subsequent stimulation is delivered, thereby ensuring the independence of evoked responses and reducing carry-over effects~\cite{DBLP:journals/tbe/RuaroBT05}.

To implement this protocol on the 3Brain high-density multi-electrode array (HD-MEA) platform, we developed a custom Python script using the official API provided by the manufacturer. This custom script enables precise control over the stimulation sequence, including the randomization of input order, the timing of pulse delivery, and the selection of specific electrode subsets for stimulation, ensuring reproducibility and flexibility in our experimental pipeline.

\subsection{Biological Network Readout}
For each stimulus presentation, neural activity is continuously recorded from the MEA for a time window extending from $2 \si{\second}$ before to $2 \si{\second}$ after the stimulus onset. To extract spiking activity from the raw extracellular signals in real time, we employed a spike detection algorithm based on a double-threshold strategy, which enables online identification of spike times.

The algorithm operates using three parameters: a sliding time window $w_s$, a low detection threshold $thr_{l}$, and a high detection threshold $thr_{h}$ used for final spike identification. Initially, for each recording channel, the algorithm scans the signal within a local window of duration $w_s = \SI{2}{\milli\second}$ to identify candidate peaks exceeding $thr_{l} \times \sigma$, where $\sigma$ denotes the standard deviation of the full recorded signal.

To refine the detection, all segments of the signal that cross the low threshold are temporarily excluded, and a new standard deviation $\sigma_n$ is computed on the remaining data. Final spike times are then determined as the local maxima that exceed the updated high threshold, defined as $thr_{h} \times \sigma_n$.

Now, let $t_s$ denote the onset time of a stimulus. For each electrode $(i, j)$, we compute the spiking activity $a^C_{ij}(t_s)$ within a temporal window $W$ following the stimulus, defined as the number of spikes observed:

\begin{equation} 
    a^W_{ij}(t_s) = \sum_{t=t_s}^{t_s+C} s_{ij}(t),
\end{equation}

where $s_{ij}(t)$ is a binary variable that equals 1 if a spike is detected at time $t$, and 0 otherwise. Time is discretized according to the sampling rate of the acquisition system, which in our setup was the maximum available frequency $f_s = 20{,}000$ Hz. We use $a^W_{ij}(t_s)$ with a properly defined time window ($W$) as the readout of the stimulus. The result is a 4096-dimensional feature vector that encodes the input pattern in the latent space of the biological reservoir. To evaluate the real network’s ability to propagate information beyond the stimulation site, a square region surrounding the stimulated electrodes is excluded from this feature vector. 

\subsection{Classifier Training and Testing Phase}

The training phase of the single-layer perceptron classifier begins with the selection of a set of input patterns, each of which is mapped onto a predefined subset of MEA electrodes for stimulation. For every input sample, electrical pulses are applied through the designated electrodes and repeated $N = 20$ times to generate a distribution of responses amenable to statistical analysis. This repetition is essential because the biological networks exhibit spontaneous activity \cite{fabri} and are subject to intrinsic noise and varying dynamical states at the moment of stimulation. As a result, the same input pattern elicits a range of different responses across trials.
Once all feature vectors have been collected, a linear classifier is trained to associate each high-dimensional representation with its corresponding input label. For every class, the recorded responses are randomly shuffled and split into disjoint sets for training and testing. Training is conducted using a single-layer perception (SLP) optimized via stochastic gradient descent (SGD), with the objective of minimizing the cross-entropy loss~\cite{DBLP:journals/nca/KlineB05}. No validation set is used, as no early stopping criterion is applied; this choice allows for maximal utilization of the available data for both training and evaluation. The model is trained for 1000 epochs.

To assess the generalization performance of the trained model, we employ a 5-fold cross-validation strategy. In each fold, the dataset is randomly partitioned into five equally sized subsets. Four subsets are used for training the linear classifier, while the remaining one is held out for testing. This process is repeated five times, such that each subset serves exactly once as the test set. The classifier remains fixed at the state reached at the end of training in each fold, with no additional updates or fine-tuning performed during testing.

Since the testing pipeline replicates the training setup, the statistical properties of the test features closely match those of the training features, ensuring a consistent and unbiased evaluation. For each test sample, its latent representation is fed into the trained classifier, and the resulting prediction is compared with the corresponding ground truth label. Performance is quantified using classification accuracy, defined as the ratio of correctly predicted samples to the total number of test samples. This metric provides a robust measure of the system's ability to serve as a biologically grounded reservoir computing architecture.

\section{Experimental Evaluation} 
\label{sec:exp}

\begin{figure*}[t]
    \centerline{\includegraphics[width=0.9 \linewidth]{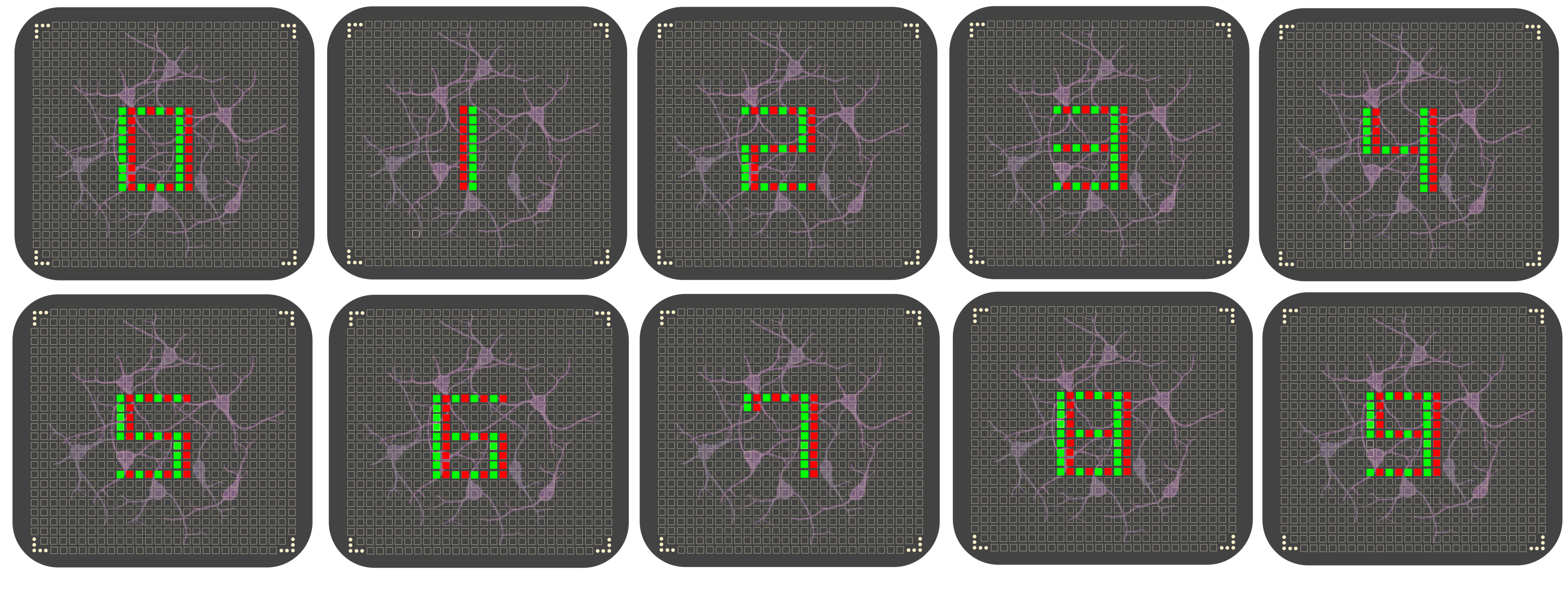}}
    \caption{\textbf{Visual representation of the input patterns.} 
    Digit recognition: input patterns represent the digits from 0 to 9.}
    \label{fig:pdf_pages}
\end{figure*}

\begin{figure}[t]
    \centerline{\includegraphics[width=1.0\linewidth]{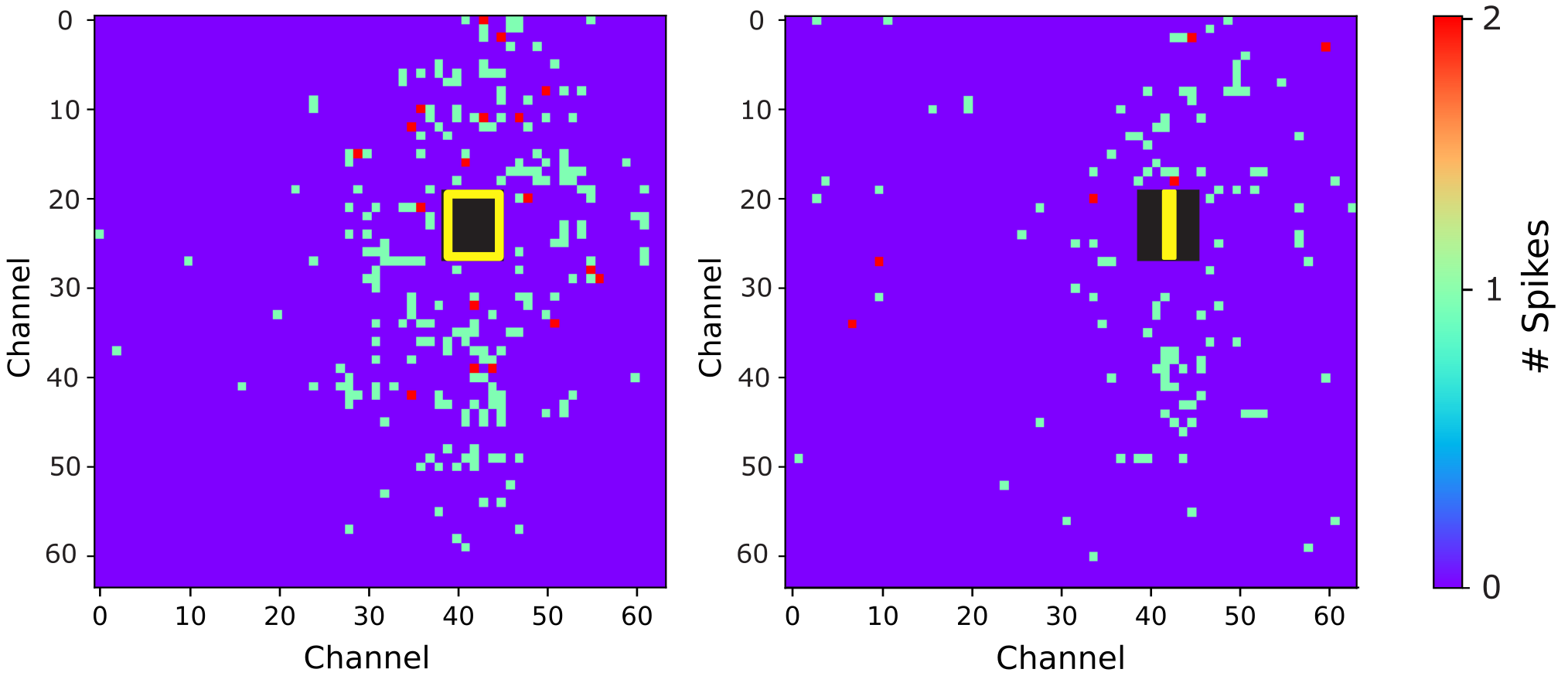}}
    \caption{\textbf{Heatmaps of neural activity following stimulation:} The left panel shows the recorded response within a $10\,\si{\milli\second}$ time window following stimulation with the input pattern "\textit{0}", while the right panel depicts the corresponding response for the input pattern "\textit{1}". The heatmaps illustrate the spatial distribution of spiking activity across the MEA electrodes.}
    \label{fig:MEA_responses}
\end{figure}

This section presents the experimental setup used to evaluate our Biological Reservoir Computing (BRC) system and discusses the results obtained using a custom digit dataset. Specifically, we assess the classification performance of the latent representations generated by the biological reservoir through three biological replicates of the same neural culture. The stimulus patterns are constructed following an \textit{electronic clock} layout. This approach emulates the segments of a digital display, where each active electrode pair represents a lit LED segment forming part of a digit.
\cref{fig:pdf_pages} provides a visual overview of the stimuli employed and \cref{fig:MEA_responses} provides two examples of the recorded responses, computed over a time window of $W = 10\,\si{\milli\second}$.

\subsection{Artificial Reservoir Comparison}
To establish a reference point and enable direct comparison, we implemented an artificial reservoir (AR) based on a network consisting of 4,096 recurrent units with a connection sparsity of 10\%—i.e., only 10\% of all possible interconnections between units were active. The AR was initialized in a resting state, with all unit activations set to zero, and driven by the same input stimuli used in the biological experiments to ensure comparability of conditions.

To replicate the variability inherent in spontaneous neural activity, we incorporated biologically realistic input noise into the AR model. This noise was empirically estimated by analyzing $N = 100$ randomly selected time windows of spontaneous activity, with a temporal duration of $W$, recorded from the biological network. The average spike count computed from these segments was used to generate synthetic noise, which was superimposed on the input stimuli before being passed to the AR. After the presentation of each stimulus, the AR dynamics were allowed to evolve for one additional time step—excluding a square region around the stimulation units, in line with the masking procedure applied in the biological experiments—was used as input to a single-layer perceptron for classification.

While the AR model serves as a valuable performance benchmark and is expected to yield superior results due to its engineered structure, it does not possess the physical or energetic characteristics of a real biological system. Consequently, the AR provides an upper bound for performance evaluation, whereas the BRC system offers insights into biologically grounded computation and energy-efficient information processing.

\subsection{Quantitative Results}
To evaluate the performance of the BRC system, it is first essential to determine the most informative time window for reading out the network's response to stimulation. For this purpose, we analyzed the classification accuracy across different stimulation sessions while systematically varying the duration $W$ of the post-stimulus time window used to extract neural activity.

\begin{figure}[t]
\centerline{\includegraphics[width=1.0\linewidth]{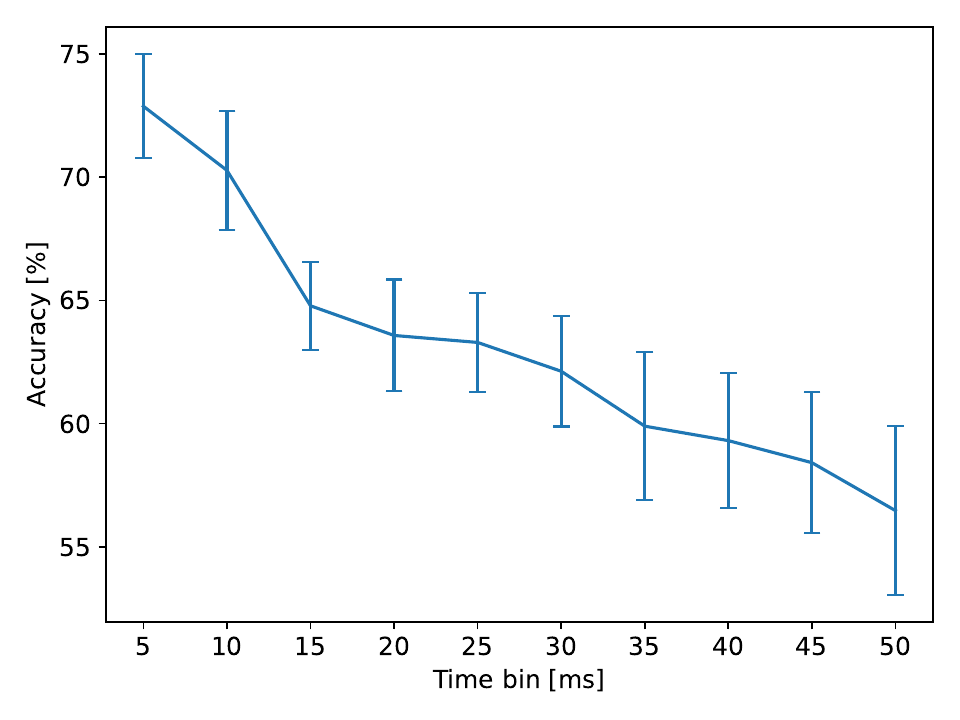}}
    \caption{\textbf{Accuracy variation across different readout windows.} For each stimulation session (n=9), classification accuracy was computed using neural responses extracted from different post-stimulus time windows, ranging from $5$ to $50\,\si{\milli\second}$.  The plot reports the mean accuracy $\pm$ standard error of the mean (SEM). This analysis highlights how the temporal integration window influences the effectiveness of the reservoir readout.}
    \label{fig:acc_var_w}
\end{figure}

\cref{fig:acc_var_w} shows the results obtained by aggregating data from $n = 9$ independent stimulation sessions performed across multiple days and biological replicates. A clear decreasing trend emerges, with the highest classification accuracy consistently observed when $W = 5\,\si{\milli\second}$. This suggests that the most informative response — in terms of discriminability among different stimuli — occurs within the first few milliseconds following stimulation.

This result is expected, as the initial segment of the neural response primarily reflects the immediate, first-order activation of neurons directly connected to the stimulated electrodes. In biological terms, a $5\,\si{\milli\second}$ window roughly corresponds to the timescale of a single synaptic transmission, capturing the direct postsynaptic potentials triggered by the stimulus. At these early latencies, the signal is minimally affected by recurrent processing, spontaneous activity, or noise propagation, making it highly specific to the input pattern.
As the time window $W$ increases beyond this point, the recorded activity becomes increasingly influenced by indirect responses, recurrent network dynamics, and spontaneous background activity. 

\begin{figure*}[t]
\centerline{\includegraphics[width=1.0\linewidth]{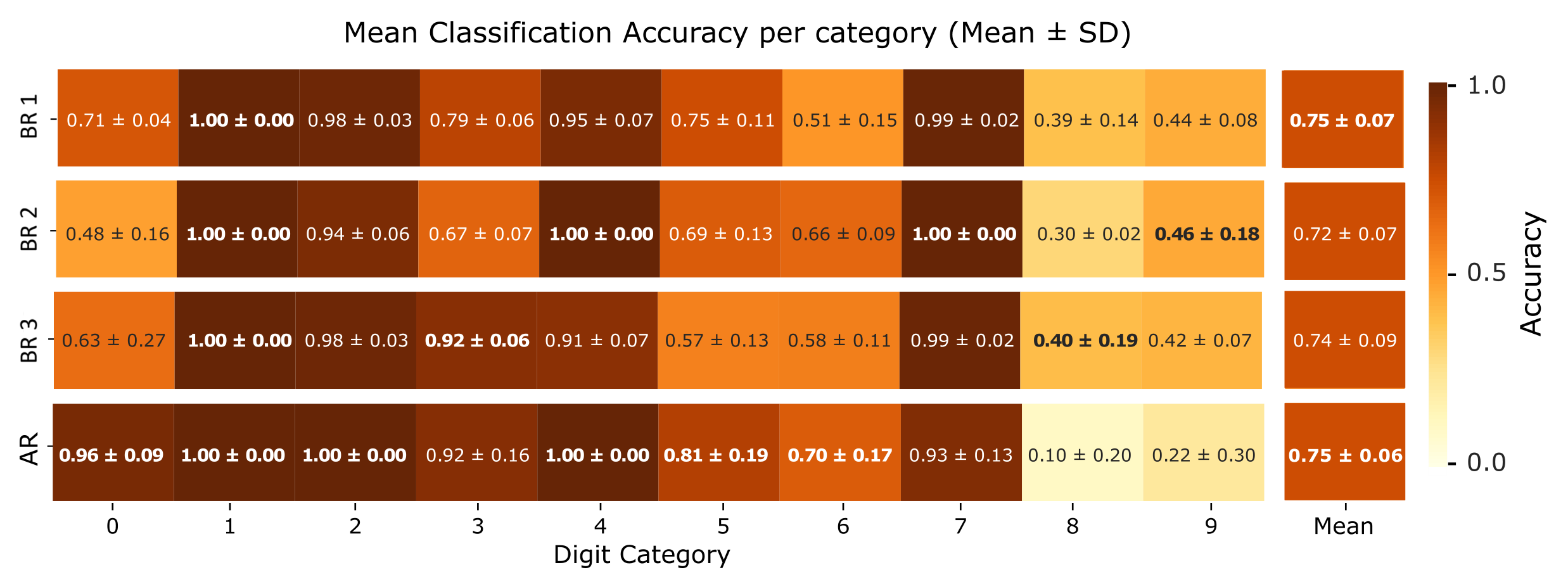}}
\caption{
\textbf{Mean classification accuracy per input category for each biological replicate.} 
Each box displays the average classification accuracy for a given stimulus category, computed over multiple stimulation sessions for each single biological replicate (BR 1, BR 2, BR 3). Accuracies are calculated using 5-fold cross-validation within each session and then averaged across the three stimulation sessions (Day 1, Day 2, Day 3).
The last row in the heatmap displays the average classification accuracy per category computed over different realizations of the noise for the Artificial Reservoir (AR).}
\label{fig:mean_acc_per_category}
\end{figure*}

\begin{figure*}[t]
\centerline{\includegraphics[width=1.0\linewidth]{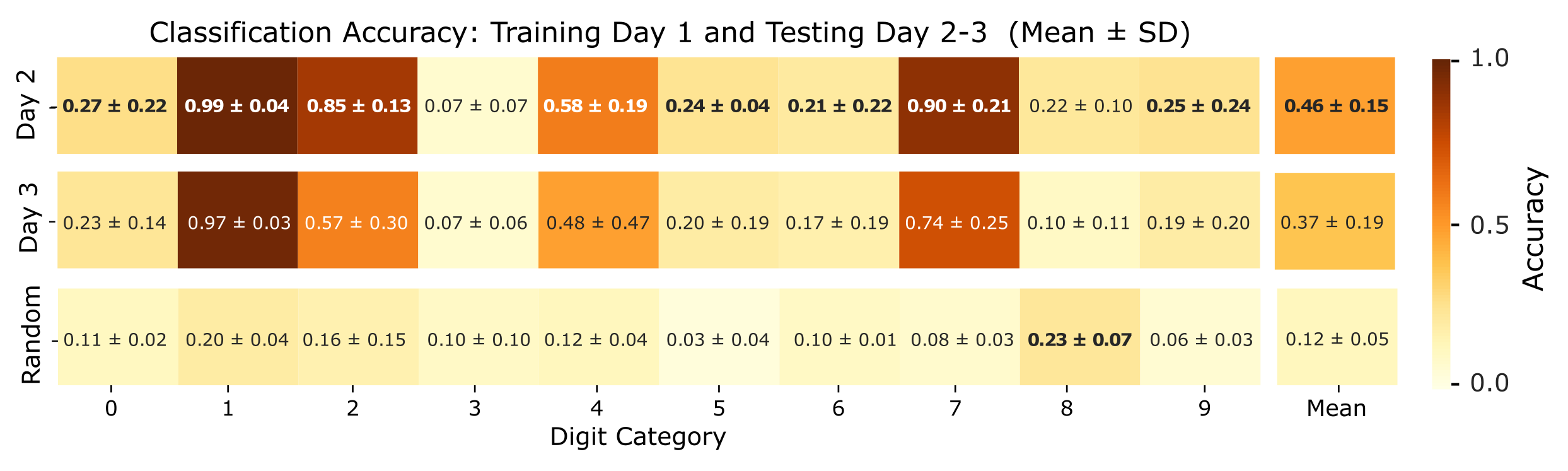}}
\caption{\textbf{Mean classification accuracy per input category across days.} 
Each box represents the average classification accuracy for a specific stimulus category, computed by training the classifier on data from Day 1 and testing on subsequent stimulation sessions (Day 2 and Day 3). Accuracies were averaged across three independent biological replicates, with each replicate evaluated using five cross-validation folds. Error bars denote the standard deviation across replicates, reflecting variability due to biological and experimental factors. The last row displays the average accuracy per category obtained for the classifier trained on Day 1 and tested on subsequent stimulation sessions by randomizing the neuronal responses for each stimulus, i.e., maintaining the absolute response but shuffling the spatial structure of the spiking activity recorded by the electrodes.}

\label{fig:acc_pretrained}
\end{figure*}

The stimulus amplitude was set to $\SI{4}{\micro\ampere}$ per electrode pair. All digit patterns were presented within the same MEA region, selected for its maximal spontaneous activity and signal quality.
The average classification accuracy per category computed over the three stimulation sessions (Day 1, Day 2, Day 3), and the comparison with the average classification accuracy of the artificial reservoir (AR) computed over different realizations of the noise are summarized in \cref{fig:mean_acc_per_category}.  The mean classification accuracy during each stimulation session across days of stimulation and biological replicates is reported in \cref{tab:acc}.

\begin{table}[t]
    \centering
    \small
    \renewcommand{\arraystretch}{1.3}
    \setlength{\tabcolsep}{6pt}
    \begin{tabularx}{\linewidth}{l|>{\columncolor{gray!10}}ccc|>{\columncolor{magenta!8}}c}
        \toprule
        \textbf{Stim. day}&\multicolumn{3}{c|}{\textbf{Biological Replicates(\%)}} & \textbf{Mean$\pm$SD(\%)} \\
        \cmidrule(lr){2-4}
        & \textbf{BR 1} & \textbf{BR 2 } & \textbf{BR 3} & \\
        \midrule
        Day 1 & 77 $\pm$ 5 & 67 $\pm$ 9 & 78 $\pm$ 6 & 74 $\pm$ 6 \\
        Day 2 & 78 $\pm$ 5 & 66 $\pm$ 9 & 73 $\pm$ 4 & 72 $\pm$ 6 \\
        Day 3 & 72 $\pm$ 6 & 68 $\pm$ 9 & 67 $\pm$ 4 & 69 $\pm$ 3 \\
        \bottomrule
    \end{tabularx}
    \caption{\textbf{Classification accuracy across stimulation day and biological replicates.} Results are reported for each stimulation session (Day 1, Day 2, Day 3) and for the three biological replicates (BR 1, BR 2, BR 3). Each value indicates the mean classification accuracy and standard deviation across 5-fold validation. The final column shows the average performance across replicates per day.}
    \label{tab:acc}
\end{table}
Interestingly, all biological reservoirs achieved performance levels comparable to those of the artificial reservoir, with an average classification accuracy of approximately 75\%. This consistency was observed across multiple stimulation sessions and among the different biological replicates, underscoring the reliability and reproducibility of the system. Notably, our biological reservoir benchmark capitalizes on the inherent advantages of using a living neural substrate, such as rich, nonlinear dynamics and natural variability, which may offer valuable insights and potential benefits over conventional artificial models.

\subsection{Accuracy Variation Across Days}
In order to assess the stability and generalizability of stimulus-evoked responses over time, we evaluated the classification performance of the classifier trained on data from the first stimulation session and subsequently tested it on data collected in the following days. This cross-day testing approach allows us to probe whether the neural dynamics elicited by a specific set of stimuli remain consistent across days, or if they are subject to spontaneous drift and reorganization.

It is well established that neural cultures evolve dynamically over time, spontaneously transitioning through various activity regimes—from uncoordinated, random spiking to highly synchronized bursting patterns~\cite{fabri, IANNELLO2025116184}. Such spontaneous reorganization is driven by intrinsic developmental processes and changes in connectivity, and it can significantly influence how the network responds to external stimulation. As a consequence, the evoked responses to identical stimulation patterns may vary substantially across days, reflecting underlying shifts in the network’s functional state. 

Then, we used all the neuronal responses to the different stimulus patterns recorded during the first stimulation session (Day 1) to train a linear classifier. The trained model was then evaluated on the data acquired during the subsequent independent stimulation sessions (Day 2 and Day 3), allowing us to assess the temporal stability and generalizability of the biological reservoir's encoding capabilities. The average classification accuracy per category, computed across the three biological replicates, is summarized in \cref{fig:acc_pretrained}. As an additional control, we assessed the classifier's performance on a surrogate dataset in which each neuronal response from the test days was randomized across recording channels. More in detail, we randomized the neuronal responses for each stimulus, maintaining the absolute response but shuffling the spatial structure of the spiking activity recorded by the electrodes. As expected, the classifier failed to discriminate among the stimulus classes under these conditions, yielding a mean accuracy of $12 \pm 5\%$, which is consistent with the chance level expected for a 10-class problem. The results revealed a progressive decrease in classification performance across days, with the accuracy dropping to $46 \pm 15\%$ on Day 2 and further to $37 \pm 19\%$ on Day 3. Although these values remain above chance level (10\%), they indicate a substantial decline in the network's ability to preserve stimulus-specific encoding over time. This performance degradation suggests that the underlying neural dynamics and connectivity patterns within the biological reservoir evolve significantly across days, likely due to spontaneous activity and plasticity phenomena intrinsic to neuronal cultures.

Interestingly, two stimulus categories—specifically, patterns "1" and "7"—consistently maintained higher classification accuracy across all sessions. While the precise reason for this robustness remains unclear, one possible explanation is that these patterns involve a lower number of stimulation electrodes and exhibit minimal spatial overlap with other stimuli. Such properties may lead to more distinct and less noisy evoked responses, facilitating their identification. However, further investigation is required to rigorously assess the relationship between stimulus complexity, spatial interference, and classification stability.

\section{Conclusion}
\label{sec:conclusions}
In this work, we proposed a novel approach to reservoir computing (RC) by leveraging a network of cultured biological neurons as the computational substrate. Unlike traditional artificial reservoirs, our bio-hybrid system exploits the intrinsic dynamics of living neurons, which operate with remarkable energy efficiency and rich nonlinear behavior. These properties make biological reservoirs a compelling alternative for neuromorphic computing, particularly in scenarios where low-power, adaptive computation is desirable. Furthermore, this line of research contributes to a deeper understanding of how biological neural circuits process information, offering insights into the mechanisms underlying cognition and neural computation.

To validate our approach, we designed and implemented a custom stimulation and recording protocol using high-density multi-electrode array (HD-MEA) technology. Our experiments focused on a digit classification task involving 10 input categories (digits 0 through 9), each mapped to a specific spatial pattern of electrical stimulation. The recorded neuronal responses were used to train a simple linear readout layer, and the classification results demonstrated that the biological reservoir was capable of generating high-dimensional feature embeddings sufficient to support accurate pattern recognition. These findings provide strong evidence that cultured neural networks can be repurposed as effective computational modules for RC applications, even in non-trivial tasks such as handwritten digit classification.

While the current study focused on a single benchmark dataset to assess system performance, future work will aim to generalize these findings by testing the BRC system on more diverse and complex input patterns. This will help evaluate the scalability and generalization capability of the biological reservoir. 

Overall, this work lays the foundation for a new class of biologically grounded reservoir computing architectures. By uniting experimental neuroscience with machine learning principles, we move closer to realizing energy-efficient, adaptive systems that bridge the gap between artificial and biological intelligence.

\section*{Acknowledgements}
This work has been supported by: Matteo Caleo Foundation; Scuola Normale Superiore (FC); PRIN project ``AICult'' (grant \#2022M95RC7) from the Italian Ministry of University and Research (MUR) (FC); PNRR project ``Tuscany Health Ecosystem - THE'' (CUP B83C22003930001) funded by the European Union - NextGenerationEU.

{
    \small
    \bibliographystyle{ieeenat_fullname}
    \bibliography{main}
}

\end{document}